\documentclass{article}

\usepackage[arxiv]{colm2026_conference}
\setcitestyle{numbers,square}

\usepackage{microtype}
\usepackage{graphicx}
\usepackage{booktabs}
\usepackage{caption}
\usepackage{hyperref}
\usepackage{xurl}       %
\usepackage{lineno}

\usepackage{amsmath,amsthm,amssymb}
\usepackage{bm}         %
\usepackage{siunitx}

\usepackage[dvipsnames,table]{xcolor}
\usepackage{multirow}
\usepackage{algorithm}
\usepackage[noend]{algpseudocode}
\usepackage{enumitem}
\usepackage{wrapfig}
\usepackage[most]{tcolorbox}

\usepackage[capitalize,noabbrev]{cleveref}

\hypersetup{
   breaklinks=true,
   pdftitle={Meta-Harness: End-To-End Optimization of Model Harnesses},
   urlcolor=NavyBlue,
   linkcolor=NavyBlue,
   citecolor=ForestGreen!70!black
}
\colorlet{my-red}{BrickRed!90!Sepia}
\colorlet{my-blue}{Aquamarine!30!Blue}
\definecolor{ours-accent}{HTML}{E45748}
\newcommand{\ours}[1]{\texttt{#1}}
\newcommand{\filename}[1]{\nolinkurl{#1}}

\tcbset{
  takeawaybox/.style={
    width=\linewidth,
    boxrule=0.8pt,
    top=6pt,
    bottom=2pt,
    left=3pt,
    right=3pt,
    before skip=1pt,
    after skip=1pt,
    colback=my-blue!3!white,
    colframe=my-blue,
    colbacktitle=my-blue,
    enhanced,
    breakable,
    attach boxed title to top left={yshift=-0.1in,xshift=0.15in},
    boxed title style={boxrule=0pt,colframe=white,},
  }
}
\newtcolorbox{takeawaybox}[2][]{takeawaybox,title=#2,#1}

\tcbset{
  logbox/.style={
    width=\linewidth,
    top=4pt,
    bottom=4pt,
    left=6pt,
    right=6pt,
    colback=black!2!white,
    colframe=black!30!white,
    fontupper=\small\ttfamily,
    before upper={\setlength{\parskip}{6pt}\setlength{\parindent}{0pt}},
    breakable,
    enhanced,
  }
}
\newtcolorbox{logbox}[1][]{logbox,#1}

\crefname{appendix}{appendix}{appendices}
\Crefname{appendix}{Appendix}{Appendices}

\usepackage{amsmath,amsfonts,bm}

\def\eqref#1{equation~\ref{#1}}

\def\1{\bm{1}}

\DeclareMathOperator*{\argmax}{arg\,max}

\title{Meta-Harness: End-to-End Optimization of Model Harnesses}

\author{Yoonho Lee \\ Stanford
\And Roshen Nair \\ Stanford
\And Qizheng Zhang \\ Stanford
\And Kangwook Lee \\ KRAFTON
\AND Omar Khattab \\ MIT
\And Chelsea Finn \\ Stanford
}

\begin{document}

\ifcolmsubmission
\linenumbers
\fi

\maketitle

\ifcolmsubmission\else
\vspace{-5mm}
\begin{center}
\footnotesize
\textbf{Project page w/ interactive demo}: \url{https://yoonholee.com/meta-harness/} \\
\textbf{Optimized harness}: \url{https://github.com/stanford-iris-lab/meta-harness-tbench2-artifact}
\end{center}
\vspace{-0mm}
\fi

\begin{figure}[h!]
\centering
\vspace{-2mm}
\includegraphics[width=0.485\linewidth]{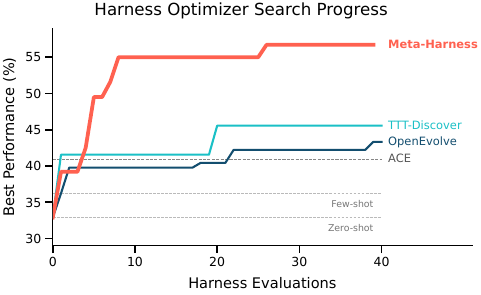}
\hfill
\includegraphics[width=0.485\linewidth]{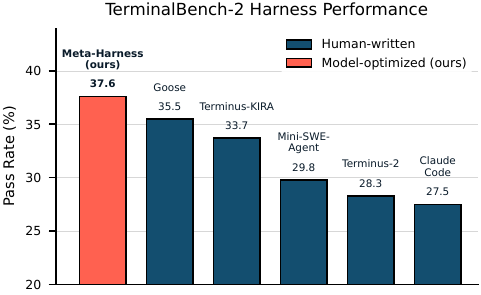}
\vspace{-2mm}
\caption{\textbf{(Left)} On text classification, Meta-Harness outperforms the best prior hand-designed harnesses (ACE) and existing text optimizers (TTT-Discover, OpenEvolve), matching the next-best method's final accuracy after just 4 evaluations. \textbf{(Right)} On TerminalBench-2, Meta-Harness outperforms all reported \texttt{Claude Haiku 4.5} harnesses.}
\label{fig:learning-curves}
\end{figure}

\begin{abstract}
The performance of large language model (LLM) systems depends not only on model weights, but also on their \textit{harness}: the code that determines what information to store, retrieve, and present to the model.
Yet harnesses are still designed largely by hand, and existing text optimizers are poorly matched to this setting because they compress feedback too aggressively: they are memoryless, condition only on scalar scores, or restrict feedback to short templates or summaries.
We introduce \textbf{Meta-Harness}, an outer-loop system that searches over harness code for LLM applications.
It uses an agentic proposer that accesses the source code, scores, and execution traces of all prior candidates through a filesystem.
On online text classification, Meta-Harness improves over a state-of-the-art context management system by 7.7 points while using 4$\times$ fewer context tokens.
On retrieval-augmented math reasoning, a single discovered harness improves accuracy on 200 IMO-level problems by 4.7 points on average across five held-out models.
On agentic coding, discovered harnesses surpass the best hand-engineered baselines on TerminalBench-2.
Together, these results show that richer access to prior experience can enable automated harness engineering.
\end{abstract}

\section{Introduction}
\label{sec:introduction}

Changing the harness around a fixed large language model (LLM) can produce a 6$\times$ performance gap on the same benchmark~\citep{Tian2026SWEBenchMC}.
The \textit{harness}---the code that determines what to store, retrieve, and show to the model---often matters as much as the model itself.
This sensitivity has led to growing interest in \textbf{harness engineering}, the practice of refining the code around an LLM to improve the overall system's performance~\citep {openai2026harnessengineering,young2025harness,boluk2026harnessproblem,bockeler2026harnessengineering}.
But despite its importance, harness engineering remains largely manual: practitioners inspect failures, adjust heuristics, and iterate on a small number of designs.
In this paper, we ask whether this process itself can be automated.

A natural starting point is recent work on text optimization, since harness engineering also involves iteratively improving text and code artifacts using feedback from prior attempts~\citep{pryzant2023protegi,romera2024mathematical,novikov2025alphaevolve,lee2025feedback,agrawal2025gepa}.
However, these methods are poorly matched to harness engineering because they typically operate with short-horizon or heavily compressed feedback: some condition only on the current candidate~\citep{madaan2023self,yang2023large,yuksekgonul2024textgradautomaticdifferentiationtext}, others rely primarily on scalar scores~\citep{novikov2025alphaevolve,cemri2026adaevolve}, and others restrict feedback to short templates or LLM-generated summaries~\citep{agrawal2025gepa,lee2025feedback}.
This is a pragmatic scalability choice, not evidence that longer-range dependencies are uninformative.
Harnesses act over long horizons: a single choice about what to store, when to retrieve it, or how to present it can affect behavior many reasoning steps later.
Compressed feedback often removes the information needed to trace downstream failures to earlier harness decisions.
Across the tasks studied by several representative text optimizers, the available context per optimization step ranges from only 100 to 30{,}000 tokens~(\Cref{tab:task_comparison}), far below the diagnostic footprint of harness search.
More broadly, work on retrieval and memory-augmented language models suggests that useful context should often be accessed adaptively rather than monolithically packed into a single prompt~\citep{lewis2020retrieval,trivedi2023interleavingretrievalchainofthoughtreasoning,packer2023memgpt,zhang2026recursivelanguagemodels}.

\begin{figure}[t]
\centering
\vspace{-11mm}
\includegraphics[width=0.94\linewidth]{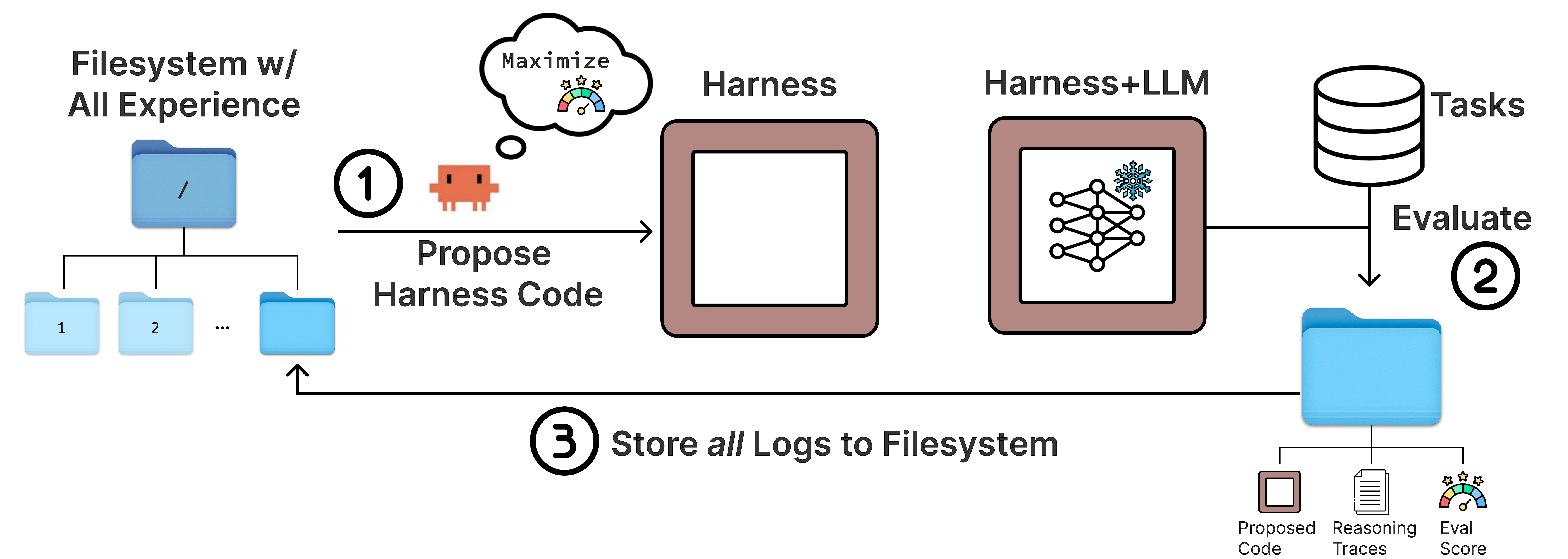}
\vspace{-3mm}
\caption{
\label{fig:method}
  \textbf{Meta-Harness search loop.}
  \textbf{(1)} An agent reads a filesystem containing all prior candidates' source code, execution traces, and scores, and proposes a new harness.
  \textbf{(2)} We evaluate the proposed harness on evaluation tasks.
  \textbf{(3)} All logs (proposed code, reasoning traces, evaluation scores) are stored in the filesystem in a new directory, and the loop repeats.
}
\end{figure}
\begin{table}[t]
\vspace{-2mm}
\centering
\footnotesize
\setlength{\tabcolsep}{4pt}
\begin{tabular}{lp{1.8cm}p{5.2cm}S[table-format=2.3,detect-weight=true]}
\toprule
Method & History & Log content & \multicolumn{1}{c}{MTok/iter} \\
\midrule
OPRO~\citep{yang2023large}            & Window & past (solution, score) pairs               & 0.002 \\
TextGrad~\citep{yuksekgonul2024textgradautomaticdifferentiationtext}        & Last & textual feedback on current artifact & 0.015 \\
AlphaEvolve~\citep{novikov2025alphaevolve}     & Window & program database + eval.\ scores           & 0.022 \\
GEPA~\citep{agrawal2025gepa}            & Summary & reflective feedback from rollout traces & 0.008 \\
Feedback Descent~\citep{lee2025feedback} & Summary & comparison + textual feedback & 0.012 \\
TTT-Discover~\citep{yuksekgonul2026learningdiscovertesttime}    & Window & prev.\ solution fragment & 0.026 \\
\midrule
\ours{Meta-Harness}   & \textbf{Full} & \textbf{\textit{all}} logs and scores & 10.0 \\
\bottomrule
\end{tabular}
\vspace{-1mm}
\caption{
\textbf{Comparison of text optimization methods and their settings.}
Each row represents a method collapsed across tasks. 
\texttt{Mtok/iter} is our best estimate of the full context generated from one evaluation of a text artifact in the \textit{largest setting considered in each paper}. 
This paper considers settings that yield orders-of-magnitude more context per artifact evaluation.}
\label{tab:task_comparison}
\vspace{-3mm}
\end{table}

We address this limitation with \textbf{Meta-Harness}, an agentic harness for optimizing harnesses via end-to-end search~(\Cref{fig:method}).
Its proposer is a coding agent, i.e., a language-model-based system that can invoke developer tools and modify code.
The choice of coding agent (rather than raw LLM) matters because the amount of experience quickly exceeds context limits, so the proposer must decide \textit{what} to inspect and validate edits through direct interaction with the codebase.
Its key design choice is to expose \textbf{full history} through a \textit{filesystem}, enabling selective diagnosis of raw prior code and execution traces rather than optimization from compressed per-candidate summaries.
For every previous candidate harness, the filesystem stores the source code, evaluation scores, and execution traces, which the proposer retrieves via standard operations such as \texttt{grep} and \texttt{cat} rather than ingesting them as a single prompt.
In practice, the proposer reads a median of \textbf{82 files per iteration} in our most demanding setting, referencing over 20 prior candidates per step~(\Cref{app:proposer-stats}).
In the settings we study, a single evaluation can produce up to 10,000,000 tokens of diagnostic information, roughly three orders of magnitude beyond the largest feedback budgets used in prior text optimization settings~(\Cref{tab:task_comparison}).

We evaluate Meta-Harness on online text classification, mathematical reasoning, and agentic coding.
On online text classification, harnesses discovered by Meta-Harness improve over Agentic Context Engineering (ACE, \citet{zhang2025ace}) by \textbf{7.7 points} while using 4$\times$ fewer context tokens, and match the next-best text optimizer's final performance after $60$ proposals with only four~(\cref{fig:learning-curves}).
On retrieval-augmented math reasoning, a single discovered harness improves accuracy on 200 IMO-level problems by \textbf{4.7 points} on average across five held-out models.
On TerminalBench-2, the discovered harness \textbf{surpasses Terminus-KIRA and ranks \#1 among all Haiku 4.5 agents}.

\section{Related Work}
\label{sec:related-work}

At a high level, Meta-Harness brings ideas from the broader literature on credit assignment and meta-learning~\citep{schmidhuber1993self,thrun1998learning,andrychowicz2016learning,finn2017maml,snell2017prototypical,akyürek2023learningalgorithmincontextlearning} in a new regime enabled by recent advances in coding agents.
Rather than updating model weights, the system assigns credit at the harness level: it uses experience from past rollouts to deliberately reason about which steps and components are responsible for failures, then rewrites the external code that governs future behavior.
More specifically, the method lies at the intersection of several recent research threads; it is most directly related to work on adaptive access to external context, executable code search, and text optimization.

\textbf{External memory and adaptive access.}
Several prior works note the benefits of treating large knowledge sources or long inputs as external resources that a language model accesses adaptively, rather than consuming them in a single pass.
Specifically, retrieval-augmented generation~\citep{lewis2020retrieval}, interleaved retrieval and reasoning~\citep{trivedi2023interleavingretrievalchainofthoughtreasoning}, memory-based agents~\citep{packer2023memgpt}, or recursive language models~\citep{zhang2026recursivelanguagemodels} are mechanisms for adaptive access to external context.
Meta-Harness uses a similar access pattern, but in the more demanding setting of harness engineering, where the proposer selectively inspects a large external history of code, scores, and execution traces to improve context-management procedures themselves.

\textbf{Executable code search.} Recent methods search over executable code for functions, workflows, or agent designs.
Early work proposes using large models as mutation and crossover operators in evolutionary program search~\citep{lehman2022evolutionlargemodels}.
Later methods evolve designated functions within fixed program scaffolds~\citep{romera2024mathematical}, use meta-agents to program new agents from prior discoveries~\citep{hu2025automated}, or search over workflow graphs for agentic systems~\citep{zhang2025aflowautomatingagenticworkflow}.
Another line of work searches over memory designs for continual-learning agents, where memory persists across task streams~\citep{zhang2025memevolve,Xiong2026LearningTC}.
In contrast, Meta-Harness searches over domain-specific harnesses, including prompt construction, retrieval, and state update strategies that reset between tasks.
Its outer loop is deliberately minimal: instead of relying on a fixed scaffold, an archive of prior discoveries, or a persistent memory mechanism, it gives the proposer unrestricted filesystem access to prior experience.
This lets the agent decide what information to inspect and enables search over full harness implementations rather than a predefined space of context-management procedures.

\textbf{Text optimization methods.} Meta-Harness is also closely related to methods such as ProTeGi, TextGrad, OPRO, GEPA, AlphaEvolve/OpenEvolve, and Feedback Descent, which iteratively improve prompts or other text artifacts using feedback from prior attempts~\citep{pryzant2023protegi,madaan2023self,yuksekgonul2024textgradautomaticdifferentiationtext,yang2023large,agrawal2025gepa,novikov2025alphaevolve,openevolve,lee2025feedback}.
However, these methods are less well suited to harness engineering, where optimization targets a complete executable procedure, and the relevant environmental feedback is distributed across code, scores, and execution traces in a way that is hard to summarize up front.
Rather than reacting only to aggregate scores or summaries, the proposer in Meta-Harness can reason over failed examples and their execution traces to propose targeted edits.
See~\Cref{tab:task_comparison} for a comparison of problem scale considered in those papers and ours, and~\Cref{fig:learning-curves,fig:learning-curves-full} for a direct comparison with OpenEvolve, GEPA, and TTT-Discover in our problem setting.

\section{Meta-Harness: A Harness for Optimizing Harnesses}
\label{sec:meta-harness}

This section describes Meta-Harness, our outer-loop procedure for searching over task-specific harnesses.
Meta-Harness is built on the idea that harness optimization benefits from allowing a proposer to selectively inspect prior code and execution traces via filesystem access, rather than optimizing from lossy summaries or an additional hand-designed search structure.
At a high level, it repeatedly proposes, evaluates, and logs new harnesses.

Meta-Harness is itself a harness in the broad sense (hence the name), since it determines what information the proposer model sees during search.
Unless otherwise noted, we use \emph{harness} to refer to the task-specific programs being optimized.

\textbf{Objective.}
A harness is a stateful program that wraps a language model and determines what context the model sees at each step.
The goal is simple: find the harness that makes the underlying model perform best on the target task distribution.
Formally, let $M$ denote a fixed language model and $\mathcal{X}$ a task distribution.
For a harness $H$ and task instance $x \sim \mathcal{X}$, we execute a rollout trajectory $\tau \sim p_M(H, x)$.
The harness constructs prompts for $M$, the model responds, and the harness updates its state after each interaction.
A task-specific reward function $r(\tau, x)$ scores the trajectory.
The objective of harness optimization is to \textbf{find the harness that maximizes the expected final reward}:
\[
H^* = \argmax_{H} \mathbb{E}_{x \sim \mathcal{X}, \tau \sim p_M(H, x)} \; r(\tau, x),
\]
When multiple objectives are relevant (e.g., accuracy and context cost), we evaluate candidates under Pareto dominance and report the resulting frontier.
In practice, this search has traditionally been carried out by human engineers and researchers, who iteratively refine prompts, context-management rules, and tool-use logic by hand.

\textbf{Meta-Harness search loop.}
Meta-Harness uses a single coding-agent proposer with access to a growing filesystem $\mathcal{D}$ that serves as its feedback channel\footnote{Based on earlier exploration, we think this workflow only became practical recently, following major improvements in coding-agent capabilities around early 2026.}.
Here, a \textit{coding agent} is a language-model-based system that can invoke developer tools and modify code.
Unlike prior systems that externalize the improvement logic in a hand-designed search loop, Meta-Harness delegates diagnosis and proposal to the coding agent itself: it decides which prior artifacts to inspect, which failure modes to address, and whether to make a local edit or a more substantial rewrite.
Equivalently, the proposer is not a raw next-token model operating on a fixed prompt assembled by the outer loop; it is an agent that retrieves information, navigates prior artifacts, and edits code as part of the search itself.
Each evaluated harness contributes a directory containing its source code, scores, and execution traces (such as prompts, tool calls, model outputs, and state updates).
The filesystem is typically far larger than the proposer's context window, so the proposer queries it through terminal tools such as \texttt{grep} and \texttt{cat} rather than ingesting it as a single prompt.
At each iteration, the proposer first inspects prior code, scores, and execution traces, then reasons about likely failure modes before generating a new harness.

Meta-Harness maintains a population $\mathcal{H}$ and a Pareto frontier over evaluated harnesses, but imposes no parent-selection rule: the proposer is free to inspect \textit{any} prior harness and its execution trace when proposing new ones.
We run evolution for a fixed number of iterations and perform a final test-set evaluation on the Pareto frontier. This simplicity is deliberate: by leaving diagnosis and edit decisions to the proposer rather than hard-coding search heuristics, Meta-Harness can improve automatically as coding agents become more capable.
The proposer never sees test-set results; its only feedback comes from the \textbf{search set}, the subset of task instances used to evaluate candidate harnesses during search and generate the feedback signal for improvement, and from execution traces logged during those search runs.

\textbf{Advantages of code-space search.}
Harness optimization occurs in code space, where small changes to retrieval, memory, or prompt-construction logic can affect behavior many steps later, making local search heuristics poorly matched to the problem.
By inspecting execution traces, the proposer can often infer \emph{why} a harness failed and which earlier design choices likely contributed to the failure, not just \emph{that} it failed, as illustrated by the search trajectories in \Cref{app:proposer-stats,app:qualitative-behavior}. There, we see that the proposer reads broadly across prior code and logs, then uses those traces to identify confounded edits, isolate likely causal changes, and shift toward safer modifications after repeated regressions.
The proposer can therefore modify the harness at the level of algorithmic structure, ranging from changes to retrieval, memory, or prompt-construction logic to full program rewrites, rather than filling in templates or applying predefined mutation operators.
In practice, it often starts from a strong prior harness, but this is an emergent strategy rather than a hard-coded rule.
Although the search space is large, representing harnesses as programs provides a natural regularization bias: coding models tend to propose coherent algorithms rather than brittle, hard-coded solutions, which biases the search toward reusable context-management procedures.
This action space is closely aligned with the read--write--execute workflows on which frontier coding assistants are trained.

\textbf{Practical implementation.}
In our experiments, each harness is a single-file Python program that modifies task-specific prompting, retrieval, memory, and orchestration logic.
In our experiments, the proposer $P$ is Claude Code~\citep{anthropic2025claudecode} with \texttt{Opus-4.6}.
The proposer is guided by a minimal domain-specific skill that describes where to write new harnesses, how to inspect previous harnesses and their execution traces, and what files it can and cannot modify.
The base model $M$ varies by domain and is always frozen; see \Cref{sec:experiments} for details.
In our experiments, a typical run evaluates roughly 60 harnesses over 20 iterations.
We provide additional tips for implementing Meta-Harness in a new domain in~\Cref{app:practical-tips}.

\begin{algorithm}[t]
\caption{Meta-Harness outer loop over harnesses}
\label{alg:evolution}
\begin{algorithmic}[1]
\State \textbf{Input:} tasks $\mathcal{X}$, LLM $M$, proposer $P$, iterations $N$
\State \textbf{Initialize:} population $\mathcal{H}$ \Comment{Initial set of valid harnesses}
\State \textbf{Initialize:} filesystem $\mathcal{D} \leftarrow \emptyset$ \Comment{stores code, scores, traces}
\For{$H \in \mathcal{H}$}
  \State $E_H \leftarrow \textrm{Evaluate}(H, M, \mathcal{X})$
  \State $\mathcal{D} \leftarrow \mathcal{D} \cup \{(H, E_H)\}$
\EndFor
\For{$t = 1 \ldots N$}
  \State Proposer $P$ queries filesystem $\mathcal{D}$ \Comment{inspects prior harnesses and scores}
  \State Proposer $P$ proposes $k$ new harnesses $\{H_1,\dots,H_k\}$
  \For{$H$ in $\{H_1,\dots,H_k\}$}
    \If{$H$ passes interface validation}
      \State $\mathcal{D} \leftarrow \mathcal{D} \cup \{(H,\textsc{Evaluate}(H,M,\mathcal{X}))\}$
    \EndIf
  \EndFor
\EndFor
\State \Return Pareto frontier of harnesses stored in $\mathcal{D}$
\end{algorithmic}
\end{algorithm}

\section{Experiments}
\label{sec:experiments}

We evaluate Meta-Harness on three task domains: online text classification, math reasoning, and agentic coding.
In each domain, we compare harnesses discovered by our search against domain-appropriate baselines using the standard evaluation metric.
Please refer to each subsection for the precise experimental setup.

We compare against two main classes of methods.
\textbf{(1) Human-designed strategies}: these are hand-crafted harnesses for each domain, representing the current state of the art in context construction.
We describe these baselines in the corresponding subsections.
\textbf{(2) Program-search methods:} these methods search over candidate harnesses using feedback and reward signals, but are designed for smaller-scale settings than harness engineering.

\subsection{Online Text Classification}
\label{subsec:text-classification}

We follow the online text classification setup of \citet{zhang2025ace,ye2026meta}: an LLM receives labeled examples one at a time, updates its memory, and is evaluated on a held-out test set.
We use \texttt{GPT-OSS-120B} as the LLM text classifier, and consider the problem of designing a harness for text classification.
We use three datasets, chosen for difficulty and domain diversity:
\textbf{LawBench} (Law)~\citep{fei2024lawbench} predicts criminal charges from case descriptions (215 classes);
\textbf{Symptom2Disease} (S2D)~\citep{gretelai_symptom_to_diagnosis_2023} predicts diseases from symptom descriptions (22 classes);
and \textbf{USPTO-50k}~\citep{schneider2016s} predicts precursor reactants from product molecules (180 classes).
We initialize the search population $\mathcal{H}$ from the main baseline harnesses in this setting: zero-shot, few-shot, ACE, and MCE.
We ran 20 evolution iterations with two candidates per iteration, producing 40 candidate harnesses.

\begin{figure}[t!]
\vspace{-8mm}
\centering
\begin{minipage}[b]{0.52\linewidth}
    \centering
    \centering
\setlength{\tabcolsep}{3.5pt}
\resizebox{\linewidth}{!}{%
\begin{tabular}{
  l
  S[table-format=2.1, detect-weight]
  S[table-format=2.1, detect-weight]
  S[table-format=2.1, detect-weight]
  |
  S[table-format=2.1, detect-weight]
  S[table-format=3.1, detect-weight]
}
\toprule
& \multicolumn{3}{c|}{Datasets} & \multicolumn{2}{c}{Avg.} \\
\cmidrule(lr){2-4}\cmidrule(lr){5-6}
Harness & {USPTO} & {S2D} & {Law} & {Acc} & {Ctx $\downarrow$} \\
\midrule
Zero-Shot & 12.0 & 63.2 & 7.0 & 27.4 & 0 \\
Few-Shot ($8$) & 14.0 & 67.9 & 21.0 & 34.3 & 2.0 \\
Few-Shot ($32$) & 13.0 & 72.2 & 21.0 & 35.4 & 7.9 \\
Few-Shot (all) & 15.0 & 78.3 & 29.0 & 40.8 & 12.3 \\
MCE~\citep{ye2026meta}\textsuperscript{\dag} & 14.0 & 83.0 & 23.0 & 40.0 & 28.5 \\
ACE~\citep{zhang2025ace}\textsuperscript{\dag} & {\bfseries 16.0} & 77.8 & 29.0 & 40.9 & 50.8 \\
\midrule
\ours{Meta-Harness} & 14.0 & {\bfseries 86.8} & {\bfseries 45.0} & {\bfseries 48.6} & 11.4 \\
\bottomrule
\end{tabular}
}
\captionof{table}{
\label{tab:main_results}
Test-set metrics for all harnesses on the three datasets.
Ctx denotes additional input tokens in context (thousands).
\dag: implementation from \citet{ye2026meta}.
$\downarrow$: lower is better.
\textbf{Meta-Harness improves online text classification accuracy while using a smaller input context.}
}

\end{minipage}
\hfill
\begin{minipage}[b]{0.46\linewidth}
    \centering
    \includegraphics[width=0.995\linewidth]{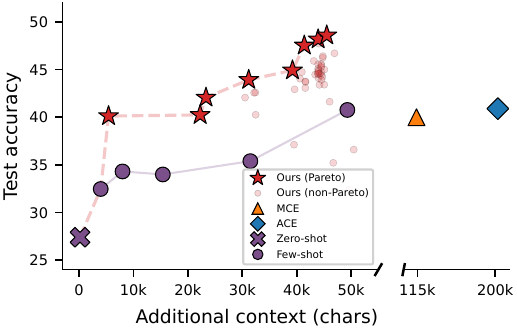}
    \vspace{-5mm}
    \caption{
    Pareto frontier of accuracy vs.\ context tokens on online text classification.
    \textbf{Meta-Harness achieves a stronger accuracy-context Pareto frontier than all comparison methods.}
    }
    \label{fig:pareto}
\end{minipage}
\end{figure}

\textbf{Comparison vs text optimizers.}
We compare Meta-Harness against representative methods for optimizing text.
For a fair comparison, we use the same proposer configuration (\texttt{Opus-4.6} with max reasoning), select candidates solely based on search-set performance, and hold out the test sets until the final evaluation.
Since evaluation is the main computational bottleneck, we give each method the same budget of proposal harness evaluations.
We consider the following points of comparison:
\begin{itemize}[leftmargin=*,noitemsep,topsep=0pt]
  \item \textbf{Best-of-N}: independent samples from the seed with no search structure; a compute-matched control for whether search matters at all.
  \item \textbf{OpenEvolve}~\citep{openevolve}: evolutionary search over programs with LLM mutation.
  \item \textbf{TTT-Discover}~\citep{yuksekgonul2026learning}: we use only the text-optimization component of their method, i.e., proposal selection via the PUCT reuse rule.
\end{itemize}
\textbf{In this setting, Meta-Harness matches the best prior text optimizers (OpenEvolve, TTT-Discover) in $0.1\times$ the evaluations}, and its final accuracy surpasses theirs by more than 10 points (\Cref{fig:learning-curves,tab:text_classification_optimizer_comparison}).
We attribute this speedup to the intentional design choices that impose minimum necessary structure on the outer loop~(\cref{sec:meta-harness}).
In particular, Meta-Harness preserves \textit{full experience history using a filesystem} and allows the proposer to inspect anything necessary, whereas both OpenEvolve and TTT-Discover operate with more structured and substantially more limited proposer inputs than full filesystem access.
We note that online text classification is the smallest-context setting we study (\Cref{tab:task_comparison}), so if structure-heavy text optimizers already lag here, their limitations may only grow in harder regimes.

\vspace{2mm}
\begin{takeawaybox}{Meta-Harness is 10$\times$ Faster and Converges to a Better Harness}
In this setting, Meta-Harness matches the best prior text optimizers (OpenEvolve, TTT-Discover) with $10\times$ fewer full evaluations, and its final accuracy surpasses theirs by more than 10 points.
\end{takeawaybox}
\vspace{2mm}

\begin{table}[t]
\vspace{-2mm}
\centering
\small
\newcommand{\xno}{\textcolor{my-red}{\texttimes}}
\newcommand{\cyes}{\textcolor{ForestGreen}{\checkmark}}
\resizebox{\columnwidth}{!}{
\begin{tabular}{lcccccccc}
\toprule
Method & Scores & Code & Summ. & Traces & Median $\uparrow$ & Best Acc $\uparrow$ & $>$ ZS \\
\midrule
Scores Only & \cyes & \cyes & \xno & \xno & 34.6 & 41.3 & 26 \\
Scores + Summary & \cyes & \cyes & \cyes & \xno & 34.9 & 38.7 & 23 \\
\midrule
\ours{Meta-Harness} (full) & \cyes & \cyes & - & \cyes & \textbf{50.0} & \textbf{56.7} & 39 \\
\bottomrule
\end{tabular}
}
\vspace{-1mm}
\caption{
\label{tab:text_classification_history_ablation}
Ablation of the information available to the proposer in online text classification.
$>$ ZS: number of runs whose accuracy exceeded the zero-shot baseline.
The full Meta-Harness interface substantially outperforms scores-only and scores-plus-summary ablations.
\textbf{Access to raw execution traces is the key ingredient for enabling harness search.}
}
\vspace{-1mm}
\end{table}

To isolate which parts of the proposer interface matter most, we compare three conditions in online text classification: a scores-only condition, a scores-plus-summary condition in which the proposer receives LLM-generated summaries but no raw traces, and the full \ours{Meta-Harness} interface with access to execution traces (\Cref{tab:text_classification_history_ablation}). The results show a large gap in favor of the full interface: scores-only reaches 34.6 median and 41.3 best accuracy, while scores-plus-summary reaches 34.9 median and 38.7 best. By contrast, \ours{Meta-Harness} reaches 50.0 median and 56.7 best accuracy, and even its median candidate outperforms the best candidate found under either ablation. We interpret this as evidence that full access to execution traces is the most important component of the interface: summaries do not recover the missing signal, and may even hurt by compressing away diagnostically useful details.

\begin{wraptable}{r}{0.38\textwidth}
\vspace{-3mm}
\centering
\small
\begin{tabular}{@{}lcc@{}}
\toprule
Method & Median & Best \\
\midrule
GEPA~\citep{agrawal2025gepa} & 32.6 & 40.2 \\
Best-of-N & 34.0 & 44.2 \\
OpenEvolve~\citep{openevolve} & 39.1 & 43.3 \\
TTT-Discover~\citep{yuksekgonul2026learning} & 34.1 & 45.6 \\
\midrule
\ours{Meta-Harness} & \textbf{50.0} & \textbf{56.7} \\
\bottomrule
\end{tabular}
\vspace{-2mm}
\caption{
\label{tab:text_classification_optimizer_comparison}
Text classification accuracies of the harnesses proposed by different text optimizers (search set).
\textbf{Meta-Harness is substantially more effective at harness optimization.}
}
\vspace{-3mm}
\end{wraptable}

\textbf{Comparison vs state-of-the-art harnesses.}
Our primary points of comparison are hand-designed harnesses for this problem setting: Agentic Context Engineering (ACE, \citet{zhang2025ace}), which uses reflective memory curation to build context over time, and Meta Context Engineering (MCE, \citet{ye2026meta}), which maintains and evolves a library of natural-language skills for context construction.
As additional baselines, we evaluate zero-shot prompting and few-shot prompting with $N \in \{4, 8, 16, 32, \text{all}\}$ examples.
Results in~\Cref{tab:main_results} show that Meta-Harness improves substantially over prior hand-designed harnesses.
The selected \ours{Meta-Harness}\footnote{We slightly overload terminology for brevity: in the tables, \ours{Meta-Harness} denotes the best discovered harness, whereas elsewhere it refers to the entire harness search procedure.} reaches 48.6\% accuracy, outperforming ACE by 7.7 points and MCE by 8.6 points.
These gains do not come from using more context: \ours{Meta-Harness} uses only 11.4K context tokens, versus 50.8K for ACE and 28.5K for MCE.

\textbf{Accuracy--Context Tradeoffs.}
Because Meta-Harness performs free-form optimization over harness code, we can express a joint preference for both accuracy and context cost rather than committing to a single scalar objective in advance.
Given only the current metrics and the desired trade-off, the proposer is able to discover harnesses across a broad range of the frontier, yielding a smooth accuracy--context Pareto curve in~\Cref{fig:pareto}.
This allows us to trade additional context for higher test accuracy in a controlled way, rather than committing to a single hand-designed operating point.

\textbf{Out-of-distribution (OOD) task evaluation.}
We evaluate whether the discovered harness generalizes to entirely new datasets unseen during search.
We consider nine diverse datasets, and describe them in detail in~\Cref{app:text-classification-data}.
The selected \ours{Meta-Harness} system achieves the best average accuracy (73.1\%), outperforming ACE (70.2\%) and all few-shot baselines~(\Cref{tab:ood_results}).
Notably, we observe that naively adding more few-shot examples beyond $32$ hurts performance in $7/9$ tasks.
Meta-Harness shows the highest performance on 6/9 datasets, suggesting that the discovered harness captures generally effective strategies for text classification rather than overfitting to the specific datasets used during search.

\begin{table*}[t]
\centering \small
\vspace{-10mm}
\resizebox{\textwidth}{!}{
\begin{tabular}{lccccccccc|cc}
\toprule
Harness & SciC & FiNER & Amz5 & FPB & GoEmo & Bank77 & News & SciT & TwHate & Avg Acc & Ctx $\downarrow$ \\
\midrule
Zero-shot & 32.7 & 56.0 & 52.7 & 90.0 & 42.0 & 80.7 & 84.7 & 89.3 & 75.3 & 67.0 & - \\
Few-shot (8) & 34.0 & 63.0 & 54.0 & 90.0 & 44.0 & 82.7 & 84.7 & \textbf{91.3} & 76.7 & 68.9 & 2.2 \\
Few-shot (32) & 38.7 & 62.0 & 53.3 & 90.7 & 43.3 & \textbf{86.0} & 85.3 & 90.7 & 76.7 & 69.6 & 5.2 \\
Few-shot (all) & 35.3 & 61.0 & 50.0 & 93.3 & 42.7 & 80.7 & 84.0 & 90.0 & 76.7 & 68.2 & 7.4 \\
ACE~\citep{zhang2025ace} & 40.7 & \textbf{74.0} & 48.0 & \textbf{96.7} & 44.0 & 83.3 & 86.0 & 90.7 & 68.7 & 70.2 & 11.7 \\
\midrule
\ours{Meta-Harness} & \textbf{53.3} & 67.0 & \textbf{60.0} & 94.0 & \textbf{46.0} & 82.7 & \textbf{86.7} & \textbf{91.3} & \textbf{77.3} & \textbf{73.1} & 7.3 \\
\bottomrule
\end{tabular}
}
\vspace{-2mm}
\caption{
\label{tab:ood_results}
OOD text classification dataset evaluation.
We report test accuracy for each dataset and the average additional context tokens across all nine datasets.
\textbf{\ours{Meta-Harness} outperforms the next best method by 2.9 points on these 9 previously unseen tasks.}
}
\vspace{-0mm}
\end{table*}

\subsection{Harnesses for Retrieval-Augmented Reasoning}
\label{subsec:math-reasoning}

We study a somewhat non-standard setup for olympiad math solving: augmenting the model with the ability to retrieve examples from a large corpus.
There is a good reason to expect retrieval to help mathematical reasoning in principle, because solutions often share reusable proof patterns, so previous reasoning traces contain information that a model may be able to exploit at inference time.
Yet retrieval has not become a standard ingredient in this setting, and prior work suggests that it has been much less successful on reasoning-intensive math benchmarks than in more fact-grounded domains~\citep{shakya2026adaptiveretrievalhelpsreasoning,xiao2024rarbreasoningretrievalbenchmark,balunovic_srimatharena_2025}.
The difficulty is that naive retrieval rarely surfaces the right traces in the right form.
This suggests that success depends less on adding retrieval per se than on discovering the right retrieval policy.
Rather than hand-designing that policy, we give Meta-Harness a hard set of olympiad problems and allow the retrieval behavior itself to emerge from search.

The retrieval corpus contains $\geq$500,000 solved problems from eight open-source datasets.
We carefully deduplicated and decontaminated it against both evaluation benchmarks and the search set, confirmed that held-out problems have no exact prefix matches under our string-based filter, and manually inspected top BM25 retrievals for held-out examples (\cref{app:math-corpus}).
We use Meta-Harness to optimize a harness for 40 iterations over a 250-problem search set of Olympiad-difficulty math problems (OlympiadBench + Omni-MATH hard), producing 109 candidate retrieval harnesses.
We initialize the search population $\mathcal{H}$ from the main baseline harnesses in this setting: zero-shot, few-shot, and ACE.
We select a single harness based on search-set performance using \texttt{GPT-OSS-20B} (\Cref{app:math-retriever}).
We evaluate this harness on $200$ previously unseen IMO-level problems drawn from IMO-AnswerBench, IMO-ProofBench, and ArXivMath~\citep{imobenchmarks2025,balunovic_srimatharena_2025}.
In addition to \texttt{GPT-OSS-20B}, we evaluate the same retrieval harness on four models not seen during search: \texttt{GPT-5.4-nano}, \texttt{GPT-5.4-mini}, \texttt{Gemini-3.1-Flash-Lite}, and \texttt{Gemini-3-Flash}.
We follow the standard evaluation protocol of prior work~\citep{imobenchmarks2025} and report accuracy averaged over three samples per problem.

\textbf{Results.}
\Cref{tab:math_results} compares the discovered harness against no retrieval, dense retrieval using the separate embedding model \texttt{text-embedding-3-small}, random few-shot prompting, and BM25 retrieval.
In contrast, Meta-Harness operates entirely in code space on top of the same BM25-based lexical retrieval stack as the sparse baseline, rather than introducing an additional dense encoder.
The discovered retrieval harness outperforms the no-retrieval baseline across all five held-out models, with an average gain of \textbf{4.7 points}.
It also matches or exceeds the strongest fixed baselines on average, outperforming BM25 retrieval by 1.3 points overall, while avoiding the regressions observed with dense retrieval and random few-shot prompting across several models.

\begin{table*}[t]
\centering \small
\vspace{-3mm}
\setlength{\tabcolsep}{4pt}
\newcommand{\posdelta}[2]{#1\,{\scriptsize(\textcolor{ForestGreen!70!black}{#2})}}
\newcommand{\negdelta}[2]{#1\,{\scriptsize(\textcolor{BrickRed}{#2})}}
\begin{tabular}{l ccccc|c}
\toprule
Method & \texttt{GPT-5.4n} & \texttt{GPT-5.4m} & \texttt{Gem-3.1FL} & \texttt{Gem-3F} & \texttt{GPT-20B} & Avg. \\
\midrule
No Retriever & 23.0 & 28.8 & 28.6 & 42.6 & 47.6 & 34.1 \\
\midrule
Dense Retrieval ($k{=}1$) & \posdelta{27.1}{+4.1} & \negdelta{24.5}{-4.3} & \posdelta{31.3}{+2.7} & \negdelta{42.3}{-0.3} & \negdelta{46.9}{-0.7} & \posdelta{34.4}{+0.3} \\
Dense Retrieval ($k{=}5$) & \posdelta{31.1}{+8.1} & \negdelta{28.3}{-0.5} & \posdelta{37.1}{+8.5} & \posdelta{47.2}{+4.6} & \negdelta{46.7}{-0.9} & \posdelta{38.1}{+4.0} \\
\midrule
Random Few-shot & \posdelta{23.1}{+0.1} & \negdelta{24.5}{-4.3} & \posdelta{31.0}{+2.4} & \negdelta{40.4}{-2.2} & \negdelta{41.8}{-5.8} & \negdelta{32.2}{-1.9} \\
BM25 Retrieval & \posdelta{30.2}{+7.2} & \posdelta{29.2}{+0.4} & \posdelta{32.8}{+4.2} & \posdelta{46.6}{+4.0} & \posdelta{48.9}{+1.3} & \posdelta{37.5}{+3.4} \\
\ours{Meta-Harness} & \posdelta{31.7}{+8.7} & \posdelta{30.4}{+1.6} & \posdelta{34.9}{+6.3} & \posdelta{46.3}{+3.7} & \posdelta{50.6}{+3.0} & \posdelta{\textbf{38.8}}{+4.7} \\
\bottomrule
\end{tabular}
\vspace{-2mm}
\caption{
\label{tab:math_results}
  Retrieval-augmented math problem solving on 200 IMO-level math problems.
  We show pass@1 averaged over three samples per problem, with absolute improvement over the baseline in parentheses.
  \textbf{The discovered Meta-Harness retrieval strategy improves reasoning on these IMO-level problems across all five held-out models, with a 4.7-point average gain over no retriever.}
}
\vspace{-8mm}
\end{table*}

\vspace{2mm}
\begin{takeawaybox}{Meta-Harness Improves Reasoning on IMO-Level Math Problems}
In retrieval-augmented math reasoning, a single discovered retrieval harness transfers across five held-out models, improving accuracy by 4.7 points on average over no retrieval and yielding the strongest overall average among the compared methods.
\end{takeawaybox}

\subsection{Evaluating Agentic Coding Harnesses on TerminalBench-2}
\label{subsec:terminalbench}

TerminalBench-2~\citep{merrill2026terminal} evaluates LLM agents on 89 challenging tasks that require long-horizon, fully autonomous execution under complex dependencies, and substantial domain knowledge.
Prior work has shown that the choice agent harness has a large effect on performance on this benchmark.
We initialize search from two strong open baselines, Terminus~2~\citep{merrill2026terminal} and Terminus-KIRA~\citep{terminuskira2026}.
For this experiment, we perform search and final evaluation on the same 89-task benchmark.
We use this benchmark as a \textit{discovery problem}~\citep{yuksekgonul2026learningdiscovertesttime} in which the goal is to discover a harness configuration that improves performance on a hard, publicly contested benchmark.
This is standard practice: public writeups already describe repeated benchmark-specific harness iteration on TerminalBench itself~\citep{forgecode_benchmarks_dont_matter,warp_terminal_bench,terminuskira2026}, and the benchmark is small and expensive enough that introducing a separate split would materially weaken the search signal.
We additionally check for overfitting by manual inspection and regex-based audits for task-specific string leakage into evolved harnesses.
We note that although the resulting harness is specialized to the TerminalBench-2 regime, autonomous completion of difficult long-horizon tasks from a single instruction is a core capability, and the benchmark consists of many tasks that frontier models and heavily engineered harnesses struggle with.

\begin{wraptable}{r}{0.42\textwidth}
\vspace{-4mm}
\centering \small
\newcommand{\xno}{\textcolor{my-red}{\texttimes}}
\newcommand{\cyes}{\textcolor{ForestGreen}{\checkmark}}
\begin{tabular}{@{}lcc@{}}
\toprule
Harness & Auto & Pass (\%) \\
\midrule
\multicolumn{3}{c}{\texttt{Claude Opus 4.6}} \\[2pt]
Claude Code          & \xno & $58.0$ \\
Terminus 2           & \xno & $62.9$ \\
Mux                  & \xno & $66.5$ \\
Droid                & \xno & $69.9$ \\
TongAgents           & \xno & $71.9$ \\
MAYA-V2              & \xno & $72.1$ \\
Terminus-KIRA        & \xno & $74.7$ \\
Capy                 & \xno & $75.3$ \\
ForgeCode            & \xno & $81.8$ \\
\midrule
\ours{Meta-Harness}  & \cyes & $\mathbf{76.4}$ \\
\midrule
\multicolumn{3}{c}{\texttt{Claude Haiku 4.5}} \\[2pt]
OpenHands            & \xno & $13.9$ \\
Claude Code          & \xno & $27.5$ \\
Terminus 2           & \xno & $28.3$ \\
Mini-SWE-Agent       & \xno & $29.8$ \\
Terminus-KIRA        & \xno & $33.7$ \\
Goose                & \xno & $35.5$ \\
\midrule
\ours{Meta-Harness}  & \cyes & $\mathbf{37.6}$ \\
\bottomrule
\end{tabular}
\vspace{-1mm}
\caption{
Pass rate on TerminalBench-2.
Results or others are from the official leaderboard.
\textbf{Meta-Harness ranks \#2 among all \texttt{Opus-4.6} agents and \#1 among all \texttt{Haiku-4.5} agents on this competitive task.}
}
\label{tab:terminalbench_results}
\vspace{-4mm}
\end{wraptable}

\textbf{Results.}
We report results on the full benchmark in~\Cref{tab:terminalbench_results}, evaluated on two base models: \texttt{Claude Opus 4.6} and \texttt{Claude Haiku 4.5}.
On \texttt{Opus 4.6}, Meta-Harness discovers a harness achieving 76.4\% pass rate, surpassing the hand-engineered Terminus-KIRA (74.7\%) and ranking \#2 among all \texttt{Opus 4.6} agents on the TerminalBench-2 leaderboard.
The only higher-scoring \texttt{Opus 4.6} agent is ForgeCode (81.8\%); however, we were unable to reproduce their reported result from the publicly available code alone, suggesting their leaderboard scores depend on components beyond the published repository.
On the weaker \texttt{Haiku 4.5} model, the improvement is larger: Meta-Harness achieves 37.6\%, outperforming the next-best reported agent (Goose, 35.5\%) by 2.1 points.
TerminalBench-2 is an actively contested benchmark with multiple teams directly optimizing for it, so the fact that an automatic search method can achieve benefits at this frontier is encouraging for long-horizon text-optimization loops.

\textbf{Qualitative behavior of the proposer.}
The harness search trajectory helps explain \textit{why} Meta-Harness achieves these gains; we provide a detailed summary in~\Cref{app:proposer-stats}.
In early iterations, the proposer combined plausible structural fixes with prompt-template edits and observed that both candidates regressed.
It then explicitly hypothesized that the regressions were confounded by the shared prompt intervention, isolated the structural changes from the prompt rewrite, and ultimately pivoted toward a safer additive modification that became the best candidate in the run.
This provides qualitative evidence that \textbf{filesystem access enables the proposer to inspect prior experience in enough detail to form causal hypotheses and revise the harness accordingly.}

\begin{center}
\begin{takeawaybox}{Meta-Harness Surpasses Hand-Engineered Agents on TerminalBench-2}
On TerminalBench-2, Meta-Harness automatically discovers harnesses that surpass Terminus-KIRA on Opus 4.6 and rank \#1 among all Haiku 4.5 agents.
\end{takeawaybox}
\end{center}

\section{Discussion}
\label{sec:discussion}

Beyond outperforming existing harnesses, Meta-Harness has several practical advantages.
Discovered harnesses generalize to out-of-distribution classification datasets (\Cref{tab:ood_results}) and to unseen base models in the math setting (\Cref{tab:math_results}).
A search run completes in a few hours of wall-clock time, yet produces readable, transferable strategies that can be reused across models, including future, stronger ones.
Overfitting in code space is also more inspectable: brittle if-chains or hard-coded class mappings are visible on inspection in a way that weight-space overfitting is not.
More broadly, our results suggest that the main advantage of Meta-Harness is not just search over code, but search with \emph{selective access to prior diagnostic experience}.
The proposer is not limited to scalar rewards or fixed summaries; it can inspect raw code, execution traces, and prior failures, then use that information to form and test hypotheses about what to change.
The qualitative search trajectories in \Cref{app:qualitative-behavior} illustrate this behavior directly.

Our findings reflect a recurring pattern in machine learning~\citep{sutton2019bitter}: once a search space becomes accessible, stronger general-purpose agents can outperform hand-engineered solutions.
A natural next step for future work is to co-evolve the harness and the model weights, letting the strategy shape what the model learns and vice versa.
While we evaluate on three diverse domains, our experiments demonstrate that harness search can work with one particularly strong coding-agent proposer (Claude Code); a broader study of how the effect varies across proposer agents remains for future work.

\clearpage
\section*{Acknowledgements}
We thank KRAFTON AI for providing API credit support.
This work is supported by OpenAI, KFAS, and Schmidt Sciences AI2050.
We thank Anikait Singh and Jubayer Ibn Hamid for their valuable feedback and suggestions, and Sienna J. Lee for patiently listening to YL's half-formed thoughts during the early stages of this work.

\bibliography{citations}
\bibliographystyle{colm2026_conference}

\clearpage
\appendix

\crefalias{section}{appendix}
\crefalias{subsection}{appendix}

\begin{figure}[t]
\centering
\includegraphics[width=0.99\linewidth]{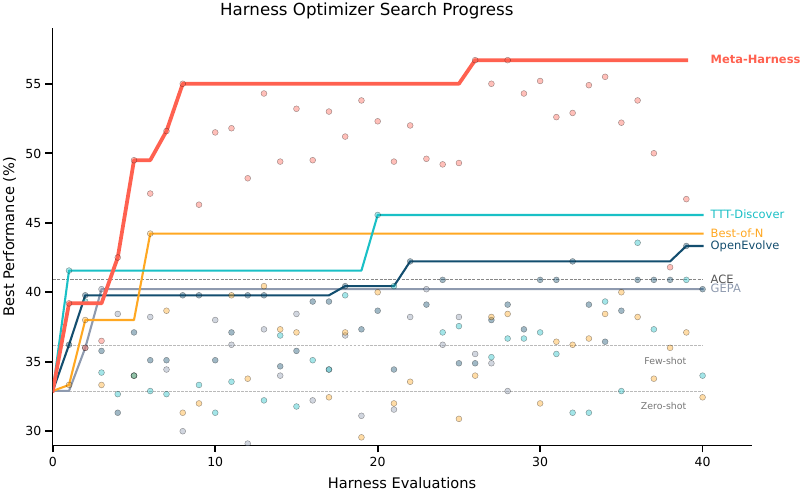}
\caption{
\label{fig:learning-curves-full}
Search-set accuracy over evaluations for all compared text optimizers on online text classification.
Each point is one candidate harness; lines track the best-so-far.
Per-dataset curves are shown alongside the aggregate.
\textbf{Meta-Harness reaches the final accuracy of OpenEvolve and TTT-Discover within the first 4 evaluations and continues improving, ending more than 10 points above all baselines.}
}
\end{figure}

\section{Qualitative Proposer Behavior}
\label{app:proposer-stats}

This section examines how the proposer uses the filesystem during search, drawing on the TerminalBench-2 run (10 iterations, \texttt{Claude Opus 4.6}).

\subsection{File Access Statistics}
\label{app:file-access-stats}

To verify that the proposer makes substantive use of the filesystem rather than defaulting to local edits, we recorded all file reads per iteration.

\Cref{tab:proposer-stats} summarizes the results.
The proposer reads a median of 82 files per iteration (range 69--99), roughly evenly split between prior harness source code (41\%) and execution traces (40\%), with the remainder going to score summaries (6\%) and other files (13\%).
This confirms that the proposer's access pattern is non-Markovian: it routinely inspects the majority of available history rather than conditioning only on the most recent parent.

\begin{table}[h]
\centering
\small
\begin{tabular}{@{}lr@{}}
\toprule
Statistic & Value \\
\midrule
Files read per iteration (median) & 82 \\
Files read per iteration (range) & 69--99 \\
\midrule
\multicolumn{2}{@{}l}{\textit{File type breakdown}} \\[2pt]
Harness source code & 41\% \\
Execution traces & 40\% \\
Score/summary files & 6\% \\
Other & 13\% \\
\bottomrule
\end{tabular}
\caption{Proposer file access statistics from the TerminalBench-2 search run (10 iterations, \texttt{Claude Opus 4.6}). The proposer reads extensively from the filesystem, with roughly equal attention to prior source code and execution traces.}
\label{tab:proposer-stats}
\end{table}

\subsection{Qualitative Behavior: Causal Reasoning Over Prior Failures}
\label{app:qualitative-behavior}

The TerminalBench-2 search log reveals a clear narrative arc in which the proposer learns from its own regressions.
Rather than wandering randomly through local edits, it forms an explicit diagnosis of why early candidates failed, then shifts toward a safer design pattern.
All text inside the log boxes below is quoted verbatim from the proposer's recorded reasoning at each iteration (emphasis ours).

\textbf{Iterations 1--2: promising bugfixes are confounded by prompt edits.}
The first two iterations both bundle plausible structural fixes with prompt-template modifications, and both regress sharply from the 64.4\% Terminus-KIRA baseline.
Iteration 1 targets observation corruption from leaked terminal markers and adds a loop breaker:

\begin{logbox}
Hypothesis: \_\_CMDEND\_\_ marker fragments leak into LLM observations on long-running tasks, causing the model to get confused and enter infinite no-tool-call loops. Stripping these markers + adding a loop breaker will recover wasted steps.
\end{logbox}

That candidate also introduced a new cleanup-oriented prompt template and a verification checklist.
Iteration 2 proposes a different state-machine fix:

\begin{logbox}
Double-confirmation completion mechanism causes verification spirals. Observed in trajectories where the agent solves the task early but burns 15--40+ additional steps re-verifying because each verification command resets \_pending\_completion, requiring another task\_complete $\rightarrow$ checklist $\rightarrow$ verify cycle.
\end{logbox}

This second candidate removes the pending-completion mechanism entirely, while also carrying over the marker stripping and the new prompt.
It still regresses, which gives the proposer two failed candidates with different structural changes but one shared prompt intervention.

\textbf{Iteration 3: the proposer identifies the confound.}
By iteration 3, the proposer explicitly infers that the regressions are not primarily due to the structural bugfixes themselves:

\begin{logbox}
Prior attempts: evo\_marker\_fix (58.9\%, -5.6pp), evo\_single\_confirm (57.8\%, -6.7pp) --- both regressed. \textbf{Root cause of regressions: Prompt template changes (cleanup directives) caused the agent to delete necessary state before task completion.} The structural bugfixes were confounded with harmful prompt changes. evo\_strip\_only isolates the two proven structural fixes.
\end{logbox}

This is the key causal step in the trajectory.
The proposer notices that the common factor across the first two failures is not the particular bugfix, but the cleanup-heavy prompt rewrite.
It therefore reverts to the original prompt and tests only the marker-stripping and loop-breaker.
The resulting candidate still underperforms slightly (63.3\%, -1.1pp), but it loses far less than the earlier versions, which supports the confound diagnosis.

\textbf{Iterations 4--6: direct fixes to the diagnosed failure mode still regress.}
The next three iterations continue to probe the same part of the design space, but now with more explicit theories about why the completion logic is fragile.
Iteration 4 attributes failures to a concrete state-machine bug in which verification commands reset the completion flag and trap the agent in repeated checklist cycles:

\begin{logbox}
Remove the two self.\_pending\_completion = False lines that reset the completion flag when intermediate commands run. This fixes a state machine bug where: (1) Agent calls task\_complete $\rightarrow$ sees QA checklist, \_pending\_completion = True (2) Agent runs verification commands $\rightarrow$ \_pending\_completion = False (bug!) (3) Agent calls task\_complete again $\rightarrow$ sees checklist AGAIN $\rightarrow$ infinite loop.
\end{logbox}

The proposer even cites concrete trajectory evidence, noting that \texttt{configure-git-webserver} produced baseline failures with agents stuck in 30--60 step verification spirals after effectively solving the task.
Iteration 5 tries to soften the cleanup language while preserving confirmation, but still edits the prompt and regresses badly.
Iteration 6 returns to the safer \texttt{evo\_strip\_only} base and proposes a systems-level optimization:

\begin{logbox}
Empty-command turns waste full LLM round-trips when terminal output hasn't changed. Smart-waiting (poll pane up to 3$\times$5s) before the next LLM call saves 5--15 turns on long-running tasks.
\end{logbox}

That change also regresses.
By this point, the proposer has learned a specific empirical lesson: modifications to prompts and completion flow are high risk, even when the local hypothesis sounds reasonable.

\textbf{Iteration 7: the winning candidate.}
After six consecutive regressions, the proposer shifts strategy from modifying the control loop to adding information before the loop begins:

\begin{logbox}
All 6 prior iterations regressed from the 64.4\% baseline because they modified the completion flow, prompt template, or observation processing. \textbf{evo\_env\_bootstrap takes a different approach --- purely additive.} It gathers an environment snapshot via a single shell command before the first LLM call and appends it to the initial prompt. No other methods are changed. This should eliminate 3--5 wasted exploration turns on dependency-heavy tasks without risking regression on already-passing tasks.
\end{logbox}

This candidate is the best result so far.
The important point is not just that iteration 7 wins, but that the proposer articulates \emph{why} it should be safer: it avoids touching the previously fragile completion machinery and instead adds information that is useful mainly on hard tasks.

\textbf{Iteration 8: composition.}
Having found one additive improvement, the proposer next attempts to compose it with an earlier structural fix:

\begin{logbox}
Combining two orthogonal fixes --- env snapshot (saves early exploration turns) + marker stripping with no-tool-call loop breaker --- will yield +1--3pp because they address independent failure modes without touching prompts or confirmation flows (which caused regressions in 5 of 7 prior iterations).
\end{logbox}

\textbf{Iteration 10: cross-run transfer.}
The proposer references results from a separate earlier search run:

\begin{logbox}
The evolution history showed ``don't cleanup service artifacts'' was worth +18pp. Iter 9 (evo\_no\_cleanup\_directive) targeted the same idea but crashed before evaluation.
\end{logbox}

\textbf{Summary.}
The search trajectory demonstrates that the proposer does more than random mutation.
Across the first seven iterations, it identifies a confound, tests the confound-isolating hypothesis directly, observes that control-flow and prompt edits remain fragile, and then deliberately pivots to a purely additive modification that becomes the best candidate in the run.
It subsequently tries to compose that winning idea with earlier fixes and even transfers lessons across runs.
This kind of causal reasoning over prior failures is precisely what full-history filesystem access enables and what compressed-feedback optimizers cannot support.

\section{Discovered Harnesses}
\label{app:discovered-harnesses}

Meta-Harness discovers executable inference-time procedures specific to the problem setup at hand.
These harnesses are structured, domain-specific policies, often with nontrivial control flow such as routing, filtering, and conditional context construction, selected solely by whether they improve search-set performance.
This section presents compact, method-style abstractions of representative harnesses that summarize the main behaviors and control-flow decisions that drive inference-time behavior.
For reference, the full implementation for each discovered harness is on the order of 100--1000 lines of code.

\subsection{Text Classification Harness}
\label{app:classification-harnesses}

In online text classification, Meta-Harness discovers a family of memory-based harnesses rather than a single canonical policy.
Table~\ref{tab:classification_pareto} reports the Pareto frontier of non-dominated variants from the main search, all selected solely by search-set performance.
We highlight two representative endpoints here: \ours{Meta-Harness (Draft Verification)}, the lowest-context frontier point, and \ours{Meta-Harness (Label-Primed Query)}, the highest-accuracy frontier point used in the main text.

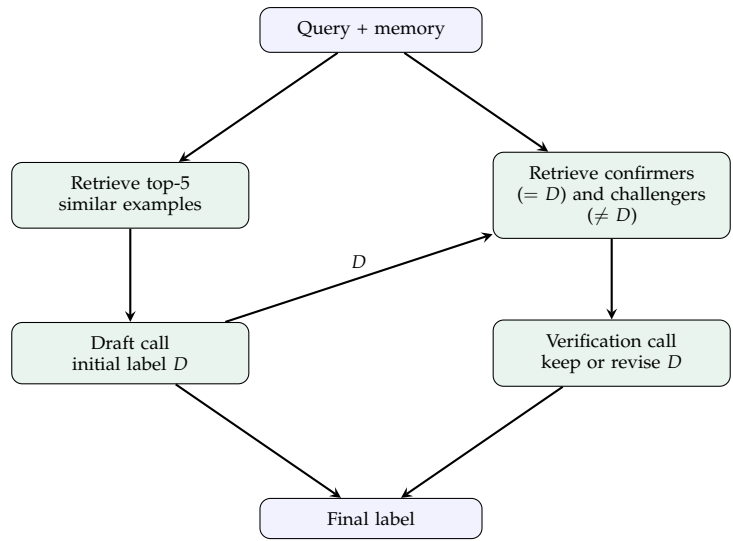
\begin{figure}[t]
\centering
\begin{tikzpicture}[>=stealth, font=\scriptsize,
    box/.style={draw, rounded corners, align=center, fill=blue!5, inner sep=5pt, text width=2.6cm},
    stage/.style={draw, rounded corners, align=center, fill=ForestGreen!8, inner sep=5pt, text width=2.8cm}]
\node[box] (query) at (0, 0) {Query + memory};
\node[stage] (draftret) at (-3.2, -2.2) {Retrieve top-5\\ similar examples};
\node[stage] (draft) at (-3.2, -4.3) {Draft call\\ initial label $D$};
\node[stage] (verifyret) at (3.2, -2.2) {Retrieve confirmers\\ ($=$ $D$) and challengers\\ ($\neq$ $D$)};
\node[stage] (verify) at (3.2, -4.3) {Verification call\\ keep or revise $D$};
\node[box] (final) at (0, -6.5) {Final label};
\draw[->, thick] (query) -- (draftret);
\draw[->, thick] (draftret) -- (draft);
\draw[->, thick] (draft) -- node[above, font=\scriptsize] {$D$} (verifyret);
\draw[->, thick] (query) -- (verifyret);
\draw[->, thick] (verifyret) -- (verify);
\draw[->, thick] (draft) -- (final);
\draw[->, thick] (verify) -- (final);
\end{tikzpicture}
\caption{\textbf{Draft-verification classification harness.}
The first call produces a draft label from a short retrieved context.
The second call retrieves evidence for and against that draft and returns the final prediction.}
\label{fig:classification-draft-verify}
\end{figure}

\begin{figure}[t]
\centering
\begin{tikzpicture}[>=stealth, font=\scriptsize,
    box/.style={draw, rounded corners, align=center, fill=blue!5, inner sep=5pt, text width=2.6cm},
    stage/.style={draw, rounded corners, align=center, fill=ForestGreen!8, inner sep=5pt, text width=2.95cm}]
\node[box] (query) at (0, 0) {Query + memory};
\node[stage] (primer) at (-3.4, -2.2) {Label primer\\ all valid labels};
\node[stage] (retrieval) at (3.4, -2.2) {TF-IDF retrieval\\ query-anchored pairing};
\node[stage] (coverage) at (-2.2, -4.6) {Coverage block\\ best example per label};
\node[stage] (contrastive) at (2.2, -4.6) {Contrastive pairs\\ similar examples\\ different labels};
\node[box, text width=5.4cm] (prompt) at (0, -7.0) {Assemble one prompt with primer, coverage, and contrastive pairs};
\node[box] (final) at (0, -9.0) {Final label};
\draw[->, thick] (query) -- (primer);
\draw[->, thick] (query) -- (retrieval);
\draw[->, thick] (retrieval) -- (coverage);
\draw[->, thick] (retrieval) -- (contrastive);
\draw[->, thick] (primer.south west) to[out=-115,in=180] ([xshift=-4pt]prompt.west);
\draw[->, thick] (coverage) -- (prompt);
\draw[->, thick] (contrastive) -- (prompt);
\draw[->, thick] (prompt) -- (final);
\end{tikzpicture}
\caption{\textbf{Label-primed query-anchored classification harness.}
The program builds a single prompt that exposes the label space, then populates it with query-relevant coverage examples and local contrastive pairs.}
\label{fig:classification-label-primed}
\end{figure}
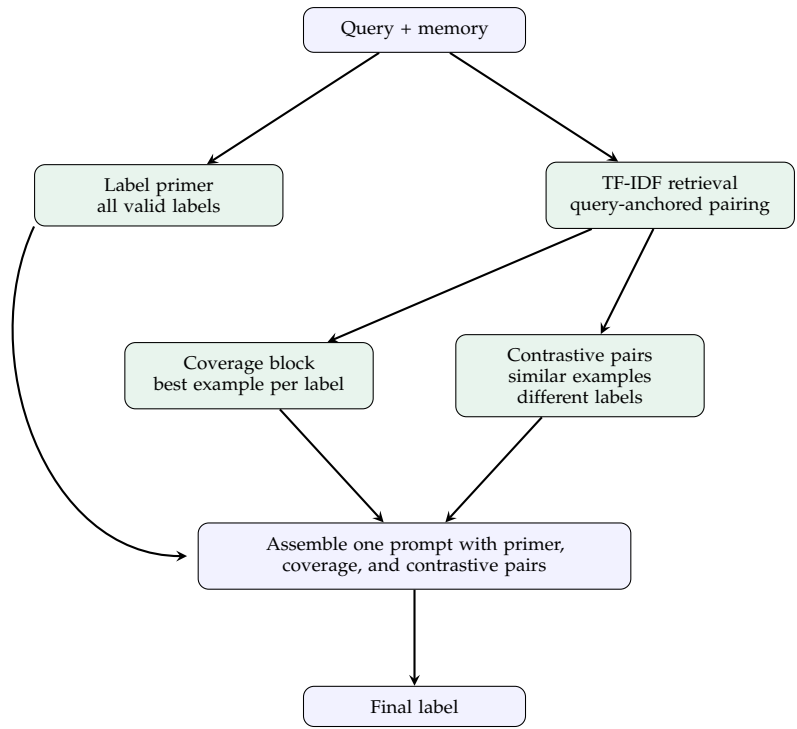

\paragraph{Overview.}
Both harnesses maintain a growing memory of past labeled examples and build prompts from that memory at inference time.
What differs is the control flow used to interrogate the memory.
\ours{Meta-Harness (Draft Verification)} uses two short calls and explicitly tests the model's first guess against retrieved counterexamples, while \ours{Meta-Harness (Label-Primed Query)} spends a larger single-call budget on making the label space and local decision boundaries explicit.
\Cref{fig:classification-draft-verify,fig:classification-label-primed} summarize these two programs.

\paragraph{Meta-Harness (Draft Verification).}
The corresponding discovered file is \filename{draft_verification.py}.
This lightweight variant turns prediction into a two-call procedure.
It first retrieves the 5 most similar labeled examples and makes a draft prediction.
It then re-queries the same memory conditioned on that draft label, retrieving 5 \emph{confirmers} with the same label and 5 \emph{challengers} with different labels, and asks the model whether to maintain or revise its initial answer.
The key discovered behavior is that the second retrieval depends on both the query and the draft prediction, so the harness can surface counterexamples targeted at the model's current guess rather than only generic near neighbors.
If too few labeled examples have been accumulated, the program falls back to a standard single-call few-shot prompt.

\begin{itemize}[leftmargin=*]
\item \textbf{Stage 1: Draft.} Retrieve the 5 nearest labeled examples and ask for an initial prediction.
\item \textbf{Stage 2: Verification.} Condition retrieval on the draft label, then show both supporting and challenging examples before making the final prediction.
\item \textbf{Cold start.} If fewer than 5 labeled examples are available, skip the two-stage procedure and use a standard single-call few-shot prompt.
\item \textbf{Why it is cheap.} Both calls use short retrieved contexts, so the overall context cost stays near the low end of the frontier even with two model invocations.
\end{itemize}

\paragraph{Meta-Harness (Label-Primed Query).}
The corresponding discovered file is \filename{label_primed_query_anchored.py}.
This strongest variant uses a single larger call built from three parts.
It begins with a \emph{label primer} listing the valid output labels, then constructs a \emph{coverage} section with one query-relevant example per label, and finally adds \emph{query-anchored contrastive pairs} that place highly similar examples with different labels side by side.
The coverage block exposes the full label space, while the contrastive block sharpens local decision boundaries around the current query.
In code, the harness implements this with TF-IDF retrieval over past labeled examples and a query-anchored pairing rule that chooses contrasting examples from the same local neighborhood.

\begin{itemize}[leftmargin=*]
\item \textbf{Label primer.} List the valid output labels before showing any examples, so the model sees the full answer space up front.
\item \textbf{Coverage block.} For each known label, retrieve the most query-relevant labeled example and include one representative example per class.
\item \textbf{Contrastive block.} Build pairs of highly similar examples with different labels, so the prompt exposes local decision boundaries around the current query.
\item \textbf{Retrieval rule.} Use TF-IDF similarity and query-anchored partner selection rather than label-agnostic nearest neighbors.
\end{itemize}
\begin{table}[t]
\centering
\resizebox{\linewidth}{!}{
\begin{tabular}{
  l
  S[table-format=2.1, detect-weight]
  S[table-format=2.1, detect-weight]
  S[table-format=2.1, detect-weight]
  |
  S[table-format=2.1, detect-weight]
  S[table-format=2.1, detect-weight]
}
\toprule
& \multicolumn{3}{c|}{Datasets} & \multicolumn{2}{c}{Avg metrics} \\
\cmidrule(lr){2-4}\cmidrule(lr){5-6}
Variant & {USPTO $\uparrow$} & {Symptom $\uparrow$} & {LawBench $\uparrow$} & {Avg $\uparrow$} & {Ctx $\downarrow$} \\
\midrule
\ours{Meta-Harness (Draft Verification)} & {\bfseries 18.0} & 85.4 & 17.0 & 40.1 & 5.4 \\
\ours{Meta-Harness (Error-Annotated)} & 9.0 & 87.7 & 24.0 & 40.2 & 22.3 \\
\ours{Meta-Harness (CoT Replay)} & 13.0 & 88.2 & 25.0 & 42.1 & 23.3 \\
\ours{Meta-Harness (Cluster Coverage)} & 12.0 & 86.8 & 33.0 & 43.9 & 31.2 \\
\ours{Meta-Harness (Cascade Retrieval)} & 12.0 & 86.8 & 36.0 & 44.9 & 39.2 \\
\ours{Meta-Harness (RRF + Contrastive)} & {\bfseries 18.0} & 89.6 & 35.0 & 47.5 & 41.4 \\
\ours{Meta-Harness (Relevance + Contrastive)} & {\bfseries 18.0} & {\bfseries 90.6} & 36.0 & 48.2 & 43.9 \\
\ours{Meta-Harness (Label-Primed Query)} & 14.0 & 86.8 & {\bfseries 45.0} & {\bfseries 48.6} & 45.5 \\
\bottomrule
\end{tabular}
}
\caption{
    Pareto-optimal discovered variants from the main text-classification search, trading off average accuracy against context cost.
    The selected system in the main text is \ours{Meta-Harness (Label-Primed Query)}.
    Ctx denotes average additional characters in input context (thousands).
}
\label{tab:classification_pareto}
\end{table}

\begin{figure}[t]
\centering
\includegraphics[width=\linewidth]{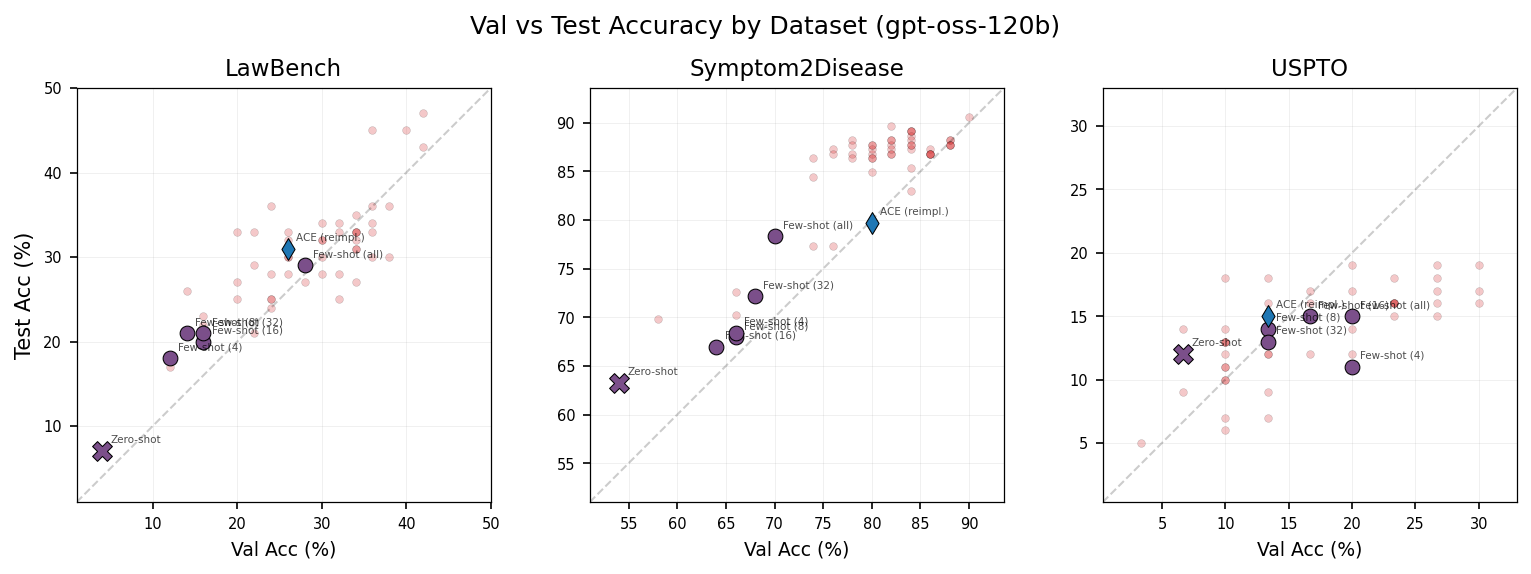}
\caption{Search-set vs.\ test accuracy per dataset for discovered text-classification strategies. Each pink dot is a discovered strategy; baselines are labeled. The dashed diagonal is $y{=}x$.}
\label{fig:val-vs-test}
\end{figure}

\subsection{Math Retrieval Harness}
\label{app:math-retriever}

This subsection describes the retrieval harness discovered by Meta-Harness for mathematical reasoning (\Cref{subsec:math-reasoning}).
The final harness is a compact four-route BM25 program whose structure emerged through search rather than being manually specified after the fact.
All design choices below---the routing predicates, reranking terms, deduplication thresholds, and per-route example counts---were selected by the outer loop across 40 iterations of evolution.

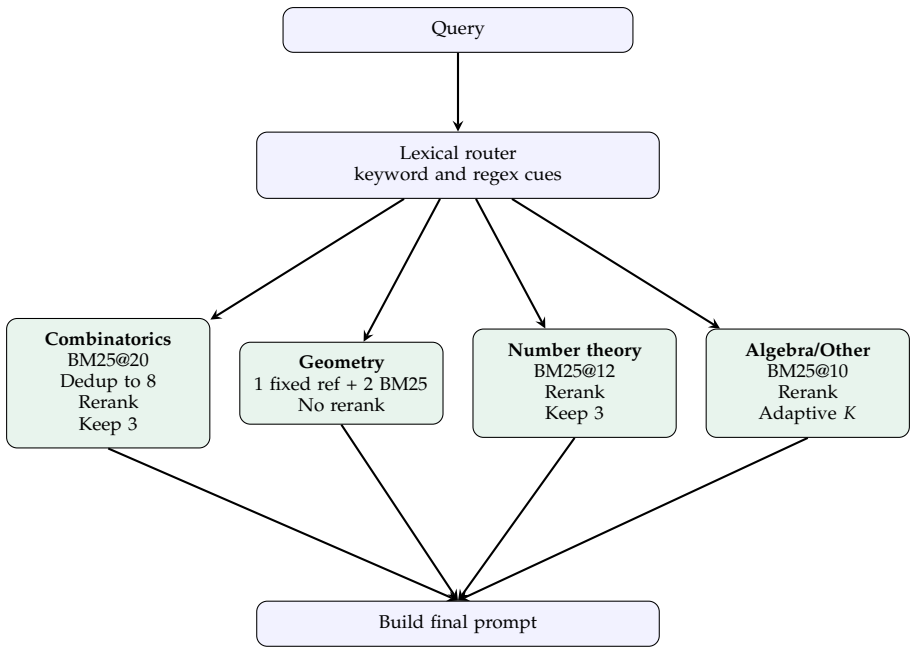
\begin{figure}[t]
\centering
\begin{tikzpicture}[>=stealth, font=\scriptsize,
    box/.style={draw, rounded corners, align=center, fill=blue!5, inner sep=5pt, text width=4.3cm},
    route/.style={draw, rounded corners, align=center, fill=ForestGreen!8, inner sep=5pt, text width=2.35cm}]
\node[box] (query) at (0, 0) {Query};
\node[box, text width=5.0cm] (gate) at (0, -1.8) {Lexical router\\ keyword and regex cues};
\node[route] (comb) at (-4.65, -4.7) {\textbf{Combinatorics}\\ BM25@20\\ Dedup to 8\\ Rerank\\ Keep 3};
\node[route] (geo) at (-1.55, -4.7) {\textbf{Geometry}\\ 1 fixed ref + 2 BM25\\ No rerank};
\node[route] (nt) at (1.55, -4.7) {\textbf{Number theory}\\ BM25@12\\ Rerank\\ Keep 3};
\node[route] (rest) at (4.65, -4.7) {\textbf{Algebra/Other}\\ BM25@10\\ Rerank\\ Adaptive $K$};
\node[box, text width=5.0cm] (prompt) at (0, -7.9) {Build final prompt};
\draw[->, thick] (query) -- (gate);
\draw[->, thick] (gate) -- (comb);
\draw[->, thick] (gate) -- (geo);
\draw[->, thick] (gate) -- (nt);
\draw[->, thick] (gate) -- (rest);
\draw[->, thick] (comb.south) -- (prompt.north);
\draw[->, thick] (geo.south) -- (prompt.north);
\draw[->, thick] (nt.south) -- (prompt.north);
\draw[->, thick] (rest.south) -- (prompt.north);
\end{tikzpicture}
\caption{\textbf{Discovered math retrieval harness.}
A lexical router assigns each query to one of four subject-specific retrieval policies.
The selected policy retrieves examples, which are inserted into the final prompt.}
\label{fig:math-harness-diagram}
\end{figure}

\paragraph{Overview.}
At inference time, the harness assigns each problem to exactly one of four routes: combinatorics, geometry, number theory, or a default route for algebra and other problems.
The gates are implemented as lightweight lexical predicates over the problem statement, including keyword sets and a small number of regex features for geometry notation.
The harness does not aggregate outputs across routes: once a route is selected, only that route retrieves examples for the final prompt.
All routes use BM25 as the underlying retrieval mechanism over the filtered corpus described above.
The BM25 index uses a math-aware tokenizer that preserves LaTeX tokens (e.g., \texttt{\textbackslash frac}, \texttt{\^{}\{2\}}) as atomic units.
The selected harness is a merge of two successful search lineages, autonomously combined by the proposer during search: one contributed a stronger geometry route based on raw BM25, while another contributed a stronger combinatorics route based on deduplication and difficulty reranking.
\Cref{fig:math-harness-diagram} gives a compact flowchart view of the final program.

\begin{itemize}[leftmargin=*]
\item \textbf{Combinatorics:} fetch 20 BM25 candidates, deduplicate to 8, rerank by lexical score and difficulty, then return the top 3. This is the main route where the harness explicitly trades off diversity against hard-problem matching.
\item \textbf{Geometry:} return 1 hard NuminaMath reference together with 2 raw BM25 neighbors. Search consistently prefers raw structural matches here over difficulty reranking.
\item \textbf{Number theory:} fetch 12 BM25 candidates and rerank using lexical score, difficulty, and a small bonus for solutions that state a technique early. This favors examples whose proof strategy is explicit.
\item \textbf{Default:} fetch 10 BM25 candidates, rerank by lexical score and difficulty, and choose an adaptive number of examples based on how concentrated the top retrieval scores are.
\end{itemize}

\subsection{TerminalBench-2 Harness}
\label{app:tbench-harness}

The discovered TerminalBench-2 harness builds on Terminus-KIRA~\citep{terminuskira2026}, inheriting its native tool calling (replacing Terminus~2's ICL-based JSON parsing), 30KB output cap, and multi-perspective completion checklist.
The main modification discovered by Meta-Harness is \textbf{environment bootstrapping}: before the agent loop begins, the harness runs a compound shell command to gather a snapshot of the sandbox environment and injects it into the initial prompt.
The proposer's hypothesis, recorded verbatim from the search log, was:

\begin{logbox}
Hypothesis: ``Injecting an environment snapshot (OS, installed languages, package managers, /app contents) before the first LLM turn will reduce wasted exploration episodes by 3--5 turns on dependency-heavy tasks''

Changes: ``Added \_gather\_env\_snapshot() that runs a single compound shell command to collect working directory, /app listing, available languages (python, gcc, node, java, rustc, go), package managers (pip, apt) [\ldots] and injects as [Environment Snapshot] block''
\end{logbox}

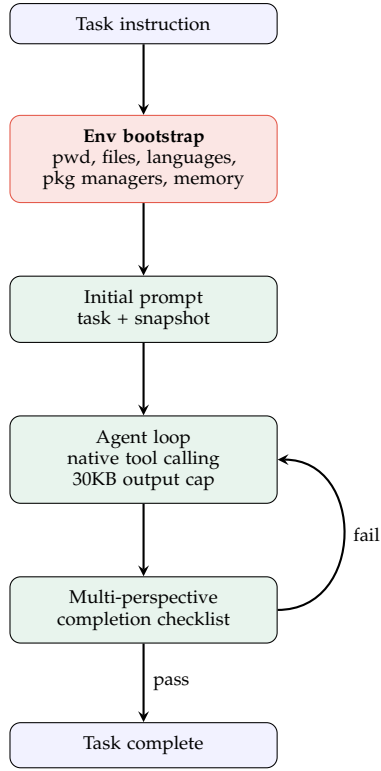
\begin{figure}[t]
\centering
\begin{tikzpicture}[>=stealth, font=\scriptsize,
    box/.style={draw, rounded corners, align=center, fill=blue!5, inner sep=5pt, text width=3.2cm},
    stage/.style={draw, rounded corners, align=center, fill=ForestGreen!8, inner sep=5pt, text width=3.2cm},
    newstage/.style={draw, rounded corners, align=center, fill=ours-accent!15, inner sep=5pt, text width=3.2cm, draw=ours-accent}]
\node[box] (task) at (0, 0) {Task instruction};
\node[newstage] (bootstrap) at (0, -1.8) {\textbf{Env bootstrap}\\ pwd, files, languages,\\ pkg managers, memory};
\node[stage] (prompt) at (0, -3.8) {Initial prompt\\ task + snapshot};
\node[stage] (loop) at (0, -5.8) {Agent loop\\ native tool calling\\ 30KB output cap};
\node[stage] (check) at (0, -7.8) {Multi-perspective\\ completion checklist};
\node[box] (done) at (0, -9.6) {Task complete};
\draw[->, thick] (task) -- (bootstrap);
\draw[->, thick] (bootstrap) -- (prompt);
\draw[->, thick] (prompt) -- (loop);
\draw[->, thick] (loop) -- (check);
\draw[->, thick] (check) -- node[right, font=\scriptsize] {pass} (done);
\draw[->, thick] (check.east) to[out=0, in=0, looseness=1.5] node[right, font=\scriptsize] {fail} (loop.east);
\end{tikzpicture}
\caption{\textbf{Discovered TerminalBench-2 harness.}
The harness inherits Terminus-KIRA's native tool calling, output cap, and completion checklist (green).
The \colorbox{ours-accent!15}{environment bootstrap} (red) is the component discovered by Meta-Harness: it gathers a sandbox snapshot before the agent loop begins, eliminating early exploratory turns.}
\label{fig:tbench-harness-diagram}
\end{figure}

The snapshot includes: the working directory, a listing of \texttt{/app} (truncated to 20 entries for large directories), available programming languages and their versions (Python, GCC, G++, Node, Java, Rust, Go), installed package managers (pip, apt-get), and available memory.
This eliminates the 2--4 exploratory turns that agents typically spend discovering what tools and files are available, allowing the model to begin productive work immediately.
The bootstrapping command is guarded by a 15-second timeout and fails silently, so it does not break the agent in unusual environments.
The full implementation adds roughly 80 lines on top of Terminus-KIRA.
\Cref{fig:tbench-harness-diagram} summarizes the harness structure.

\paragraph{Per-task analysis.}
Compared to Terminus-KIRA, the discovered harness gains on 7 of 89 tasks, with the largest improvements on \texttt{protein-assembly} and \texttt{path-tracing}.
The gaining tasks share a common property: they require domain-specific tooling whose availability cannot be assumed in advance (bioinformatics libraries, rendering pipelines, chess engines, cryptographic utilities, CoreWars simulators).
Without the bootstrap, the agent spends its first 2--4 turns probing the environment; on tasks with tight turn budgets or where early wrong assumptions cascade, those wasted turns can be the difference between pass and fail.
This suggests that the bootstrap's value is largest when the environment is non-obvious, and the task requires the agent to match its strategy to what is actually installed.

\section{Dataset Details}
\subsection{OOD Text Classification Datasets}
\label{app:text-classification-data}

\begin{itemize}[leftmargin=*,itemsep=2pt,topsep=2pt]
\item \textbf{SciCite} is a 3-way citation-intent classification benchmark introduced by \citet{cohan2019structuralscaffoldscitationintent}. Each example consists of a citation context from a scientific paper, labeled by the citation's rhetorical role, such as background, method, or result. The task tests whether a model can infer why one paper cites another from the local scientific context.
\item \textbf{FiNER-139} is a financial numeric entity recognition benchmark introduced by \citet{Loukas_2022}. It consists of word-level annotations from financial filings with 139 fine-grained XBRL entity types, making it substantially more fine-grained than standard sentence-level classification tasks. The benchmark tests whether a model can identify and classify numeric financial entities from context.
\item \textbf{Amazon Reviews} is the English portion of the Multilingual Amazon Reviews Corpus introduced by \citet{keung2020multilingualamazonreviewscorpus}. In our setting, it is used as a 5-way review rating prediction task, where the label corresponds to the review's star rating. This benchmark evaluates general-domain sentiment and rating prediction from product review text.
\item \textbf{Financial PhraseBank} is a 3-way financial sentiment benchmark introduced by \citet{malo2013gooddebtbaddebt}. It consists of sentences from financial news and related economic text labeled as positive, neutral, or negative with respect to market sentiment. The task evaluates domain-specific sentiment classification in finance.
\item \textbf{GoEmotions} is a fine-grained emotion classification benchmark introduced by \citet{demszky2020goemotionsdatasetfinegrainedemotions}. It contains English Reddit comments annotated with 27 emotion categories plus a neutral category, and is commonly treated as a 28-way classification task. The benchmark tests nuanced affect recognition beyond coarse positive-negative sentiment.
\item \textbf{Banking77} is a fine-grained intent classification benchmark introduced by \citet{casanueva2020efficientintentdetectiondual}. It contains online banking user utterances labeled with 77 intents, covering a wide range of customer service requests. The task evaluates single-domain intent detection with a large label space.
\item \textbf{AG News} is a 4-way news topic classification benchmark commonly associated with the text classification setup of \citet{zhang2016characterlevelconvolutionalnetworkstext}. Examples are labeled with broad news categories such as world, sports, business, and science/technology. It is a standard general-domain benchmark for topic classification.
\item \textbf{SciTail} is a science-domain textual entailment benchmark in which the task is to predict whether a hypothesis is entailed by a premise sentence in a science-focused inference setting~\citep{Khot_Sabharwal_Clark_2018}.
\item \textbf{TweetEval (Hate)} is the hate-speech subset of the TweetEval benchmark introduced by \citet{barbieri2020tweetevalunifiedbenchmarkcomparative}. It is a binary tweet classification task for detecting hateful versus non-hateful content within a unified social-media evaluation suite. This benchmark tests robust classification in noisy, short-form social media text.
\end{itemize}

\subsection{Math Retrieval Corpus}
\label{app:math-corpus}

\Cref{tab:math-corpus} lists the datasets composing the retrieval corpus used in \Cref{subsec:math-reasoning}.
The raw sources contain more problems than the final corpus; several filtering steps were applied before merging.
NuminaMath-1.5 was filtered to competition-math subsets (AMC/AIME, olympiad references, number theory, inequalities, and related sources), discarding lower-quality web-scraped entries.
OpenMathReasoning was deduplicated to one solution per problem (retaining the solution with the highest pass rate on an independent verifier), and problems whose source matched any evaluation benchmark family (IMO, AIME, HMMT, SMT, USAMO, Putnam) were removed before deduplication.
The entire corpus was then decontaminated against all evaluation benchmarks and the search set used during harness search, using exact prefix matching followed by fuzzy Jaccard similarity (threshold 0.8); any corpus problem matching an eval problem under either criterion was discarded.
Solutions from OpenMathReasoning and DeepMath are truncated to 5{,}000 characters to limit retrieval context length.
At runtime, the selected harness further restricts retrieval to entries with non-empty solutions shorter than 4{,}000 characters.
Retrieved solutions are truncated again to 3{,}000 characters when inserted into the prompt.
For the geometry route, the harness also constructs a separate hard-reference index from NuminaMath problems with difficulty greater than 6.

\begin{table}[h]
\centering
\begin{tabular}{lrrl}
\toprule
\textbf{Dataset} & \textbf{Problems} & \textbf{Sol. Len} & \textbf{Proof} \\
\midrule
\href{https://huggingface.co/datasets/nvidia/OpenMathReasoning}{OpenMathReasoning} & 281,743 & 5{,}000$^\dagger$ & 34\% \\
\href{https://huggingface.co/datasets/zwhe99/DeepMath-103K}{DeepMath-103K}     & 103,021 & 5{,}000$^\dagger$ & 0\%  \\
\href{https://huggingface.co/datasets/AI-MO/NuminaMath-1.5}{NuminaMath-1.5}    & 129,520 & 1{,}376           & 13\% \\
\href{https://huggingface.co/datasets/AIMO-Corpus/PolyMath}{PolyMath}          &  11,083 &     363           & 0\%  \\
\href{https://huggingface.co/datasets/KbsdJames/Omni-MATH}{Omni-MATH}         &   4,289 &     829           & 0\%  \\
\href{https://huggingface.co/datasets/SPIderman5/FineProofs-SFT}{FineProofs-SFT}    &   4,275 &   3{,}977         & 100\% \\
\href{https://huggingface.co/datasets/gneubig/aime-1983-2024}{AIME 1983--2024}   &     933 &     ---           & 0\%  \\
\href{https://huggingface.co/datasets/Putnam-AXIOM/putnam-axiom-dataset-v1}{Putnam-AXIOM}      &     492 &     888           & 100\% \\
\midrule
\textbf{Total}    & \textbf{535,356} & 5{,}000$^\dagger$ & 22\% \\
\bottomrule
\end{tabular}

\smallskip
{\footnotesize $^\dagger$ Truncated at 5{,}000 characters; actual solutions are longer.}
\caption{
    Datasets in the math retrieval corpus (535K problems total).
    Sol. \ Len is the median solution length in characters.
    Proof indicates whether the dataset contains proof-type problems (by answer or problem type field).}
\label{tab:math-corpus}
\end{table}

\subsection{Math IMO-level Test Set}
\label{app:math-per-dataset}

The main text aggregates results over 200 IMO-level problems drawn from IMO-AnswerBench, IMO-ProofBench, ArXivMath December 2025, and ArXivMath January 2026.
The 200-problem evaluation set consists of a stratified 100-problem subset of IMO-AnswerBench, together with all problems from the other three benchmarks.
This per-benchmark breakdown is useful because the four datasets mix answer-style, proof, and research-style problems, which are aggregated together in the main paper for brevity.
When included, the table in this section should report each benchmark separately for both \texttt{Base} and \texttt{Meta-Harness} across the five held-out models.

\begin{table}[h]
\centering
\begin{tabular}{lr}
\toprule
\textbf{Dataset} & \textbf{Problems} \\
\midrule
IMO-AnswerBench & 100 \\
IMO-ProofBench & 60 \\
ArXivMath Dec.\ 2025 & 17 \\
ArXivMath Jan.\ 2026 & 23 \\
\midrule
\textbf{Total} & \textbf{200} \\
\bottomrule
\end{tabular}
\caption{
    Breakdown of the 200-problem IMO-level evaluation set.}
\label{tab:math_eval_breakdown}
\end{table}

\section{Practical Implementation Tips}
\label{app:practical-tips}

Meta-Harness is largely domain-agnostic: we expect it to apply in any setting where a language model is wrapped by a task-specific harness.
Applying it in a new domain, however, requires operating in a relatively new regime of LLM-assisted coding, where the proposer conditions on long-horizon histories of prior runs and writes programs whose effects may only become visible many steps later.
In getting this workflow to work reliably, we found a small set of practical choices that mattered consistently across the three domains studied in this paper.
The guidelines below are not themselves scientific claims about the method; they are engineering lessons from building and running the system, which we hope will make it easier for future work to apply Meta-Harness in other domains.

\begin{itemize}[leftmargin=*]

\item \textbf{Write a good skill.}
The skill text is the primary interface for steering the search, and its quality is the strongest lever on whether the loop works.
The proposer receives a natural-language skill~\citep{agentskills} that defines its role, the directory layout, CLI commands, and output format.
In practice, the skill should constrain outputs and safety-relevant behavior, not the proposer's diagnosis procedure: it should specify what is forbidden, what artifacts to produce, and what objectives to optimize, while leaving the model free to inspect scores, traces, and prior code as needed.
Our intuition from inspecting logs from Meta-Harness runs is that after enough iterations, the accumulated traces often shape the proposer's behavior more than the skill itself.
In our experience, iterating on the skill text had a larger effect on search quality than changing iteration count or population size.
Expect to run a few short evolution runs (3--5 iterations each) specifically to debug and refine the skill before committing to a full run.

\item \textbf{Start with a baseline harness and a search set that is hard for it.}
Write a simple baseline (e.g., few-shot prompting), then construct the search set by either filtering for examples that the baseline gets wrong or selecting a diverse subset of difficult instances.
The search has little to optimize if the baseline already saturates the evaluation.
Keep the search set small enough for roughly 50 full evaluations per run (50--100 examples in our classification experiments, 88 problems for math retrieval); a fast, discriminative eval is more valuable than a large one.

\item \textbf{Log everything in a format that is easy to navigate.}
Evaluation code should write code, scores, and execution traces in a form that the proposer can query reliably.
In practice, this means using machine-readable formats such as JSON, organizing artifacts hierarchically, choosing reasonable and consistent file names, and adopting naming schemes that make simple tools such as regex search work well.

\item \textbf{Make logs queryable through a small CLI (optional, but helpful).}
Each harness gets a directory containing source code, scores, and execution traces, but as the history grows, raw filesystem access alone becomes cumbersome.
A short CLI that lists the Pareto frontier, shows top-$k$ harnesses, and diffs code and results between pairs of runs can make the experience store much easier to use, and querying such CLIs is closely aligned with the workflows on which coding agents are trained.
If relevant offline experience exists (rollouts from other models, solved problem corpora, relevant papers), converting it into the same directory structure can also help warm-start exploration and ground new ideas.
This layer helps the proposer save tokens it may have wasted on navigation.

\item \textbf{Lightweight validation before expensive benchmarks.}
Write a small validation test that imports the module, instantiates the class, and calls both methods on a tiny set of examples.
Harnesses proposed during the search should pass this test before being fully evaluated.
A simple test script can catch most malformed or nonfunctional candidates in seconds and keep the cost of failures near zero.

\item \textbf{Automate evaluation outside the proposer.}
Running evals is simple enough that it is not worth making the proposer do it.
A separate harness should score candidates and write results to the filesystem.

\end{itemize}

\section{Extended Related Work}
\label{app:extended-related-work}

This appendix expands the brief discussion in \Cref{sec:related-work} and situates Meta-Harness relative to several neighboring lines of work that we could not cover in detail in the main text. A recurring distinction is that Meta-
Harness optimizes executable harness implementations and provides the proposer with selective access to prior code, scores, and execution traces via the filesystem.

\paragraph{AlphaEvolve / OpenEvolve.}
AlphaEvolve~\citep{novikov2025alphaevolve} and OpenEvolve~\citep{openevolve} evolve code via LLM-guided mutations with structured feedback: the proposer receives a program database with scalar scores (4--22K tokens per step; \Cref{tab:task_comparison}) and applies fixed mutation strategies to tournament-selected parents.
These methods are designed for algorithm discovery and optimization (mathematical conjectures, scheduling heuristics, hardware kernels), where the search target is a single stateless function with a clean scalar objective, and mutations are local.
Harness engineering is a different regime: harnesses are stateful programs that accumulate experience across many examples, and a single design choice (e.g., what to store in memory) can cascade through an entire evaluation sequence.
Meta-Harness addresses this by giving an unstructured coding agent full filesystem access, letting it selectively read any prior candidate's source code, execution traces, and scores.

\paragraph{GEPA.}
GEPA~\citep{agrawal2025gepa} is the closest text optimizer in terms of feedback richness, providing rollout traces per candidate.
It is designed for prompt optimization on tasks with short feedback loops (math problems, instruction-following, code optimization), where each rollout is a single LLM call or a short pipeline.
In this regime, per-candidate reflection works well: one prompt, one answer, one score.
Harness engineering requires reasoning across many examples and many candidates simultaneously: understanding why a retrieval strategy works for one class of problems but degrades on another requires comparing execution traces across the full population.
GEPA operates on one candidate at a time (2--8K tokens per step; \Cref{tab:task_comparison}), with a fixed critique format that must anticipate what information is relevant.
Meta-Harness gives the proposer access to \textit{all} prior candidates simultaneously and lets the agent decide what to examine.

\paragraph{Prompt orchestration frameworks.}
Several systems provide structured abstractions for composing multi-stage LLM programs. LMQL~\citep{Beurer_Kellner_2023}, LangChain~\citep{chase2022langchain}, and DSPy~\citep{khattab2023dspycompilingdeclarativelanguage} make prompt engineering more systematic by providing higher-level interfaces for prompt templates, control flow, and modular LLM pipelines. These frameworks help developers specify and organize LLM programs, but they still typically require manual design of retrieval policies, memory updates, and orchestration logic. Meta-Harness operates at a different level: it searches over the \emph{implementation} of these policies in executable code, treating the harness itself as the optimization target.

\end{document}